\documentclass{article}
\usepackage[preprint]{spconf}
\usepackage{amsmath,graphicx,hyperref}
\usepackage{amssymb}
\usepackage{algorithm}
\usepackage{algpseudocode}
\usepackage{enumitem}
\usepackage{booktabs}
\usepackage{xcolor}
\usepackage{soul}
\usepackage{placeins}
\usepackage{siunitx}

\title{Compressed Active Subspaces for Scalable Bayesian Inference}

\name{Thomas Flynn$^{\star}$ \qquad Sanket Jantre$^{\star}$ \qquad Byung-Jun Yoon$^{\star \dagger}$ \qquad Kibaek Kim$^{\ddagger}$}
\address{$^{\star}$ Computing and Data Sciences Directorate, Brookhaven National Laboratory, Upton, NY, USA \\
$^{\dagger}$ Department of Electrical \& Computer Engineering, Texas A\&M University, College Station, TX, USA \\
$^{\ddagger}$ Mathematics and Computer Science Division, Argonne National Laboratory, Lemont, IL, USA
}

\copyrightnotice{\copyright This work has been submitted to the IEEE for possible publication. Copyright may be transferred without notice.}
\newcommand{\mc}{\mathcal}

\newcommand{\bs}{\boldsymbol}

\let\oldtextbf\textbf
\renewcommand{\textbf}[1]{\oldtextbf{\boldmath #1}}

\begin{document}
\maketitle
\begin{abstract}
Active subspace methods provide a framework for quantifying predictive uncertainty in high-dimensional models by identifying and performing inference along parameter directions that have the greatest influence on the model output. However, the construction of active subspaces requires storing many full-dimensional model gradients, which becomes prohibitive as model size increases. We address this limitation by proposing \emph{Compressed Active Subspaces (CAS)}, a scalable approach that first maps the model parameters to a compressed space using a structured isometric embedding and then constructs the active subspace within this reduced parameterization. Our approach substantially reduces the memory required for active subspace construction and enables Bayesian inference for large models where standard active subspace methods become impractical. We demonstrate the scalability of CAS on neural networks of increasing size while maintaining predictive performance and robust uncertainty estimates.
\end{abstract}
\vspace{-0.1in}
\section{Introduction}
\vspace{-0.05in}
\label{sec:intro}
Bayesian inference for deep learning models provides a principled framework for quantifying predictive uncertainty. However, its computational cost grows rapidly with the dimensionality of the parameter space. Active subspaces offer an efficient way to address this challenge by identifying a low-dimensional linear subspace spanned by parameter directions that have the greatest influence on the model output \cite{jantre2024active}. Recent works show that Bayesian inference restricted to such active subspaces can provide predictive uncertainty estimates comparable to those obtained from full-network Bayesian inference while remaining computationally efficient \cite{jantre2024active}. Leveraging active subspaces has been shown to enable uncertainty quantification (UQ) for relatively large deep learning models~\cite{abeer2024leveraging} and uncertainty-guided fine-tuning~\cite{abeer2026enhancing}.

A key step in the active subspace approach is identifying the dominant eigenvectors of the covariance matrix $\mc{C}$ of model gradients:
\vspace{-0.05in}
\begin{equation}
\label{eq:covariance}
    \mathcal{C} = \mathbb{E}\left[\nabla_{\boldsymbol{\theta}} f_{\boldsymbol{\theta}}(\boldsymbol{x})\nabla_{\boldsymbol{\theta}} f_{\boldsymbol{\theta}}(\boldsymbol{x})^T\right].
\end{equation}
Here, the expectation is taken over the model parameters $\theta$ and inputs $\boldsymbol{x}$. Following \cite{jantre2024active}, we consider parameter perturbations $\boldsymbol{\theta} \sim \mathcal{N}(\boldsymbol{\theta}_0,\sigma^2_0\mathbf{I})$ around a pretrained model's parameters $\boldsymbol{\theta}_0$ and sample $\boldsymbol{x}\sim p$ from the training data distribution. The eigenvectors associated with the largest eigenvalues of $\mathcal{C}$ identify parameter directions along which the model output is most sensitive to these perturbations.

In practice, the dominant eigenvectors of $\mathcal{C}$ can be approximated using a truncated SVD of sampled model gradients. Let $\theta_1,\ldots,\theta_M$ denote i.i.d. samples from $\mathcal{N}(\boldsymbol{\theta}_0,\sigma^2_0\mathbf{I})$ and let $\boldsymbol{x}_1,\ldots,\boldsymbol{x}_M$ denote samples from the training data distribution. We can construct the gradient matrix
\vspace{-0.02in}
\begin{equation}
\label{eq:gradient_matrix}
    \mathcal{G} = [\boldsymbol{g}_1,\ldots,\boldsymbol{g}_M] \in \mathbb{R}^{N\times M}, \quad \boldsymbol{g}_i = \nabla_{\boldsymbol{\theta}} f_{\boldsymbol{\theta}_i}(\boldsymbol{x}_i),
    \vspace{-0.02in}
\end{equation}
where $N$ is the number of model parameters. The leading left singular vectors of $\mathcal{G}$ provide approximate dominant eigenvectors for $\mathcal{C}$. While this approach has shown promising results for neural networks \cite{jantre2024active,abeer2024leveraging,abeer2026enhancing}, its application to large models is limited by the memory required to store $\mathcal{G}$. In particular, each additional sample of the model gradient requires storing another $N$-dimensional vector, resulting in $O(NM)$ memory.

We address this limitation by first compressing the model parameter space before constructing the active subspace. We propose \emph{Compressed Active Subspaces (CAS)}, which introduces a fixed low-dimensional parameterization of the model using an efficient isometric embedding. Active subspace construction is then performed within this compressed parameter space, substantially reducing the memory required to store the sampled model gradients. The resulting active subspace can be mapped back to the original model parameter space for quantifying predictive uncertainty using the Bayesian inference framework of \cite{jantre2024active}.

Our main contributions in this work are as follows:
\begin{itemize}[noitemsep, topsep=0pt]
    \item We propose Compressed Active Subspaces for scalable active subspace construction in models with high-dimensional parameter spaces. CAS uses a structured isometric embedding to construct active subspaces using compressed model gradients.
    \item We demonstrate the scalability of CAS on neural networks of increasing size and evaluate its predictive uncertainty estimates using active subspace-based Bayesian inference.
\end{itemize}

\noindent \textbf{Related Work.} Bayesian inference over low-dimensional subspaces provides an efficient alternative to inference over the entire neural network parameter space. Previous approaches construct low-dimensional parameter subspaces using SGD trajectories or learned regions of the neural network loss landscape \cite{izmailov2020subspace, wortsman2021learning, dold2024bayesian}. More recently, subspace inference has been combined with parameter-efficient representations to enable Bayesian inference in large language models \cite{samplawski2025scalable}. 

Active subspaces originate from the computer model literature, where gradient information is used to identify directions in input space that capture the most variability in a model output \cite{constantine2014active,constantine2015active}. Recent works extend this approach to neural network parameter spaces for uncertainty quantification in deep learning and generative models \cite{jantre2024active, abeer2024leveraging, abeer2026enhancing}. Other approaches improve the scalability of Bayesian neural networks through sparse subnetwork selection \cite{jantre2023layer, jantre2024spike, li2024ssvi}.

Randomized projections have been used for efficient computation in numerical linear algebra and machine learning. In randomized linear algebra, they form the basis of methods for low-rank approximations, randomized SVD, and preconditioning \cite{randomlin2020, Chen2020Nov, Pilanci2017Feb}. Randomized compression has also been used in distributed machine learning to reduce the communication and storage costs associated with high-dimensional model gradients \cite{compressionflynn2025, psgd, Safaryan2021Dec, Ivkin2019Dec}.

Low-rank parameterizations are widely used for parameter-efficient fine-tuning of large neural networks. Low-rank adaptation (LoRA) \cite{hu2022lora} restricts model updates to low-rank parameterizations while recent methods further reduce the trainable parameters through structured projections and parameter sharing \cite{li2025unilora}. Bayesian inference and uncertainty quantification have also been studied within LoRA parameter spaces \cite{yang2024bayesianlora, rahmati2025clora,jantre2025uncertainty}. The projection used in CAS is closely related to the isometric construction proposed in Uni-LORA \cite{li2025unilora} and to CountSketch matrices \cite{countsketch2004, countsketch2017}. In contrast to the Uni-LoRA construction, we include random signs in the nonzero entries and use the resulting embedding to enable \emph{compressed active subspace} construction.

\vspace{-0.15in}
\section{Preliminaries}
\label{sec:preliminaries}
\vspace{-0.05in}

\noindent \textbf{Bayesian subspace inference.} Let $\mathcal{D}=\{(\boldsymbol{x}_i,y_i)\}_{i=1}^n$ denote the training data and let $\boldsymbol{\theta}_0 \in \mathbb{R}^N$ denote the pretrained neural network model parameters. Bayesian inference over the full model requires inferring the posterior distribution $p(\boldsymbol{\theta}|\mathcal{D})$, which becomes expensive as $N$ grows. Subspace inference instead restricts parameter variation to a $K$-dimensional linear subspace around $\boldsymbol{\theta}_0$ \cite{jantre2024active, izmailov2020subspace}:
\vspace{-0.05in}
\begin{equation}
\label{eq:subspace}
    \boldsymbol{\theta} = \boldsymbol{\theta}_0 + \boldsymbol{V}\boldsymbol{z}, \qquad \boldsymbol{z}\in\mathbb{R}^{K},
    \vspace{-0.05in}
\end{equation}
where $\boldsymbol{V} \in \mathbb{R}^{N\times K}$ contains the basis vectors of the subspace. Bayesian inference is then performed over the reduced parameters $\boldsymbol{z}$ instead of the full model parameters $\boldsymbol{\theta}$. One can use a Gaussian prior $p(\boldsymbol{z})=\mathcal{N}(\boldsymbol{\mu}_{\boldsymbol{z}}, \tilde{\sigma}^{2}\mathbf{I}_{K})$ and obtain posterior predictive distributions via Bayesian model averaging using samples from $p(\boldsymbol{z}|\mathcal{D})$. Posterior inference over $\boldsymbol{z}$ can be performed using standard Markov chain Monte Carlo (MCMC) sampling or variational inference methods as in \cite{jantre2024active}.

\begin{figure}
\begin{minipage}{0.48\textwidth}
\begin{algorithm}[H]  
\caption{Standard Active Subspace Construction}
\begin{algorithmic}[1]
\State \textbf{input} Model $f$, weights $\boldsymbol{\theta}_0$, \# gradient samples $M$, AS dimension $K$, perturbation variance $\sigma_0^2$
\For {$m= 1,2,\dots, M$} \label{das:grad1}
\State Sample $\boldsymbol{x}_m \sim  p$
\State Sample $\boldsymbol{\theta}_m \sim \mathcal{N}(\boldsymbol{\theta}_0,\sigma_0^2\mathbf{I})$ 
\State Compute gradient $\boldsymbol{g}_m \gets \nabla_{\boldsymbol{\theta}} f_{\boldsymbol{\theta}_m}(\boldsymbol{x}_m)$
\EndFor \label{das:grad2}
\State Let $\mathcal{G}=[\boldsymbol{g}_1,\hdots,\boldsymbol{g}_M]\in\mathbb{R}^{N\times M}$
\State Compute SVD $\mathcal{G}=\boldsymbol{U}\boldsymbol{\Sigma}\boldsymbol{W}^T$
\State \Return top $K$ left singular vectors $\boldsymbol{U}_K=[\boldsymbol{u}_1,\hdots,\boldsymbol{u}_K]$
\end{algorithmic}
\label{alg:classicas}
\end{algorithm}
\end{minipage}
\vspace{-1em}
\end{figure}

\vspace{0.1in}
\noindent\textbf{Active subspace inference.}
For active subspace inference, the columns of $\boldsymbol{V}$ in \eqref{eq:subspace} are the leading left singular vectors of the sampled gradient matrix $\mathcal{G}$ in \eqref{eq:gradient_matrix}. These directions correspond to the dominant eigenvectors of the gradient covariance matrix $\mathcal{C}$ in \eqref{eq:covariance}. Algorithm~\ref{alg:classicas} summarizes the standard active subspace construction. CAS modifies this construction by first mapping the model parameters to a compressed parameter space before computing the active directions.

\vspace{-0.1in}
\section{Compressed Active Subspaces}
\label{sec:cas}
\vspace{-0.05in}

In compressed active subspaces, we introduce a fixed isometric embedding $P \in \mathbb{R}^{N\times R}$, where $R\ll N$. The reduced parameters $\boldsymbol{\phi} \in \mathbb{R}^R$ are mapped to the full model parameter space as 
\vspace{-0.05in}
\begin{equation*}
    \boldsymbol{\theta} =  \boldsymbol{\theta}_0 + \boldsymbol{P} \boldsymbol{\phi}.
\end{equation*}
We sample $\boldsymbol{P}$ randomly using the structured construction described below and keep it fixed throughout active subspace construction. We then perform active subspace construction on the reduced model $f_{\boldsymbol{\theta}_0+\boldsymbol{P}\boldsymbol{\phi}}(\boldsymbol{x})$. This reduces the dimension of each sampled model gradient from $N$ to $R$.

Our choice of $\boldsymbol{P}$ is guided by computational efficiency and storage cost. A general $N \times R$ matrix requires $O(NR)$ storage and would introduce another large matrix for models with high-dimensional parameter spaces.
We instead use a normalized CountSketch-like construction \cite{countsketch2004,countsketch2017}. 
Each model parameter is assigned independently and uniformly to one of $R$ classes. Let $M_i \in \{1,\ldots,R\}$ denote the class assigned to parameter $i$ and let $n_j$ denote the number of parameters assigned to class $j$.  
We additionally sample an independent random sign $s_i \in \{-1,+1\}$ for each parameter. The entries of $\boldsymbol{P}$ are then 
\vspace{-0.05in}
\begin{equation}
\label{eq:countsketch}
    P_{i,j} = \mathbf{1}_{j=M_i}\frac{s_i}{\sqrt{n_j}}.
    \vspace{-0.02in}
\end{equation}
Since different columns have disjoint support and each column is normalized, the resulting embedding satisfies $\boldsymbol{P}^{T}\boldsymbol{P}=\mathbf{I}_{R}$. Thus, $\boldsymbol{P}$ preserves the Euclidean geometry of the reduced parameter space when embedded in the full parameter space.

The construction in \eqref{eq:countsketch} is closely related to the isometric projection used in Uni-LoRA~\cite{li2025unilora}.  The main difference is the random sign $s_i$ applied to each nonzero entry, as in CountSketch matrices. Since each row of $\boldsymbol{P}$ contains only one nonzero entry, the embedding can be represented using the $N$ class assignments and random signs rather than storing a dense $N\times R$ matrix. Likewise, compression and decompression using multiplication by $\boldsymbol{P}$ or $\boldsymbol{P}^{T}$ can also be performed in $O(N)$ operations.

For CAS, we sample reduced perturbations as $\boldsymbol{\phi}_m \sim \mathcal{N}( \boldsymbol{0},\sigma_0^2\mathbf{I}_{R})$ and evaluate the model at $\boldsymbol{\theta}_m = \boldsymbol{\theta}_0 + \boldsymbol{P}\boldsymbol{\phi}_m$. The corresponding gradient is 
\vspace{-0.05in}
\begin{equation}
\label{eq:compressed_gradient}
    \boldsymbol{q}_m = \nabla_{\boldsymbol{\phi}} f_{\boldsymbol{\theta}_0+\boldsymbol{P}\boldsymbol{\phi}_m}(\boldsymbol{x}_m) = \boldsymbol{P}^{T}\boldsymbol{g}_m \in \mathbb{R}^{R}.
    \vspace{-0.05in}
\end{equation}
We therefore construct the compressed gradient matrix
\vspace{-0.05in}
\begin{equation*}
    \widetilde{\mathcal{G}} = [\boldsymbol{q}_1,\ldots,\boldsymbol{q}_M] \in \mathbb{R}^{R\times M},
    \vspace{-0.05in}
\end{equation*}
instead of storing the full gradient matrix $\mathcal{G} \in \mathbb{R}^{N\times M}$. The leading $K$ left singular vectors of $\widetilde{\mathcal{G}}$ define the active directions in the compressed parameter space. These directions can then be mapped back to the original model parameter space through $\boldsymbol{P}$ for Bayesian subspace inference.

Let $\mathcal{C}_{R}$ denote the gradient covariance matrix in the compressed parameter space corresponding to the embedding $\boldsymbol{P}$. From \eqref{eq:compressed_gradient},
\begin{equation}
\label{eq:compressed_covariance}
    \mathcal{C}_{R} = \mathbb{E}\left[\boldsymbol{q}\boldsymbol{q}^{T}\right] = \boldsymbol{P}^{T}\mathcal{C}\boldsymbol{P},
\end{equation} 
Therefore, CAS identifies the dominant active directions within the $R$-dimensional subspace spanned by $\boldsymbol{P}$. If $\boldsymbol{U}_{K}\in\mathbb{R}^{R\times K}$ contains the leading eigenvectors of $\mathcal{C}_{R}$, the corresponding directions in the original parameter space are $\boldsymbol{V}=\boldsymbol{P}\boldsymbol{U}_{K}$. These directions are orthonormal since $\boldsymbol{P}^T\boldsymbol{P}=\boldsymbol{I}_R$.

\begin{figure}
\begin{minipage}{0.48\textwidth}
\begin{algorithm}[H]  
\caption{Compressed Active Subspace Construction}
\label{alg:cas}
\begin{algorithmic}[1]
\State \textbf{input} Model $f$, weights $\boldsymbol{\theta}_0$,
intermediate dimension $R$, AS dimension $K$, \# gradient samples $M$, variance $\sigma_{0}^2$
\State Sample embedding
$\boldsymbol{P}\in\mathbb{R}^{N\times R}$ according to \eqref{eq:countsketch}
\State Form reduced model $\widehat{f}_{\bs{\phi}} = f_{\bs{\theta}_0 + \bs{P}\bs{\phi}}$.
\State $\boldsymbol{U}_K \gets \mathrm{StandardActiveSubspace}(\widehat{f},\boldsymbol{0},M,K,\sigma_{0}^2)$
\State \Return $\boldsymbol{V}=\boldsymbol{P}\boldsymbol{U}_K$
\end{algorithmic}
\end{algorithm}
\end{minipage}
\vspace{-1em}
\end{figure}

\vspace{-0.1in}
\section{Experiments}
\label{sec:expts}
\vspace{-0.05in}
Our experiments demonstrate the scalability of CAS on neural networks of increasing size across several datasets and evaluate its benefit for predictive uncertainty estimation using active subspace-based Bayesian inference. Except for Gaussian Process Regression, all models are neural networks with a single hidden layer, where the hidden layer size is varied to test scalability. We use Pyro for Bayesian inference \cite{pyro} with the No-U-Turn-Sampler (NUTS) throughout.

\begin{figure}[tb]
\begin{minipage}[b]{0.496\linewidth}
  \centering
  \centerline{\includegraphics[width=\linewidth]{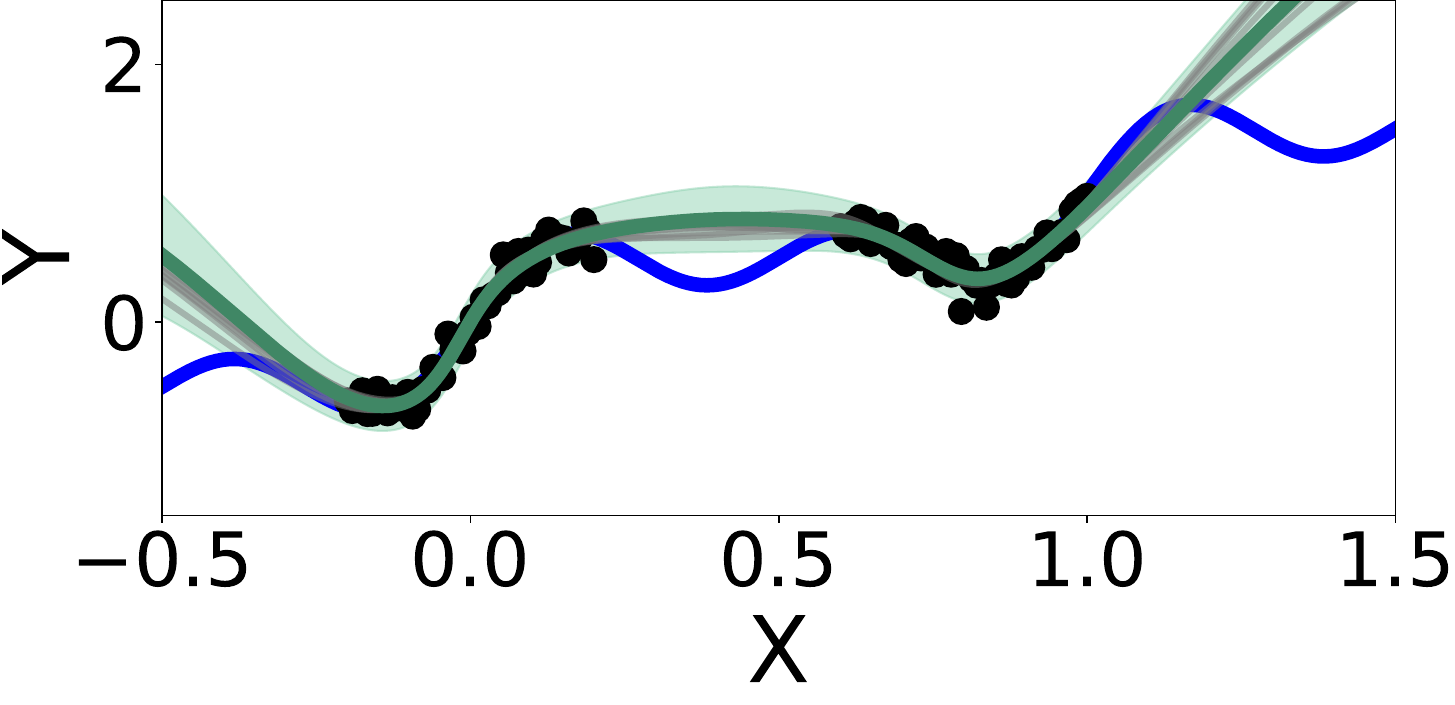}}
\end{minipage}
\begin{minipage}[b]{.496\linewidth}
  \centering
  \centerline{\includegraphics[width=\linewidth]{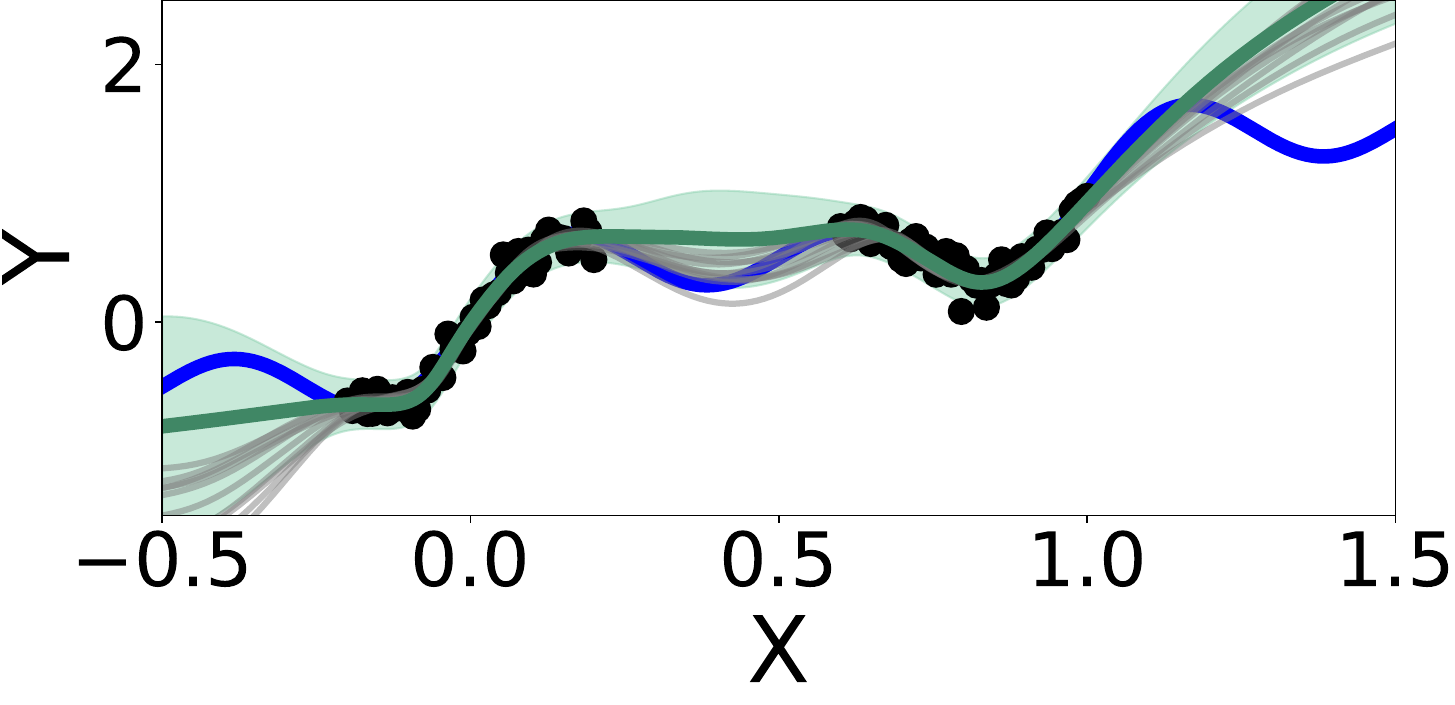}}
\end{minipage}
\hfill
\begin{minipage}[b]{0.496\linewidth}
  \centering
  \centerline{\includegraphics[width=\linewidth]{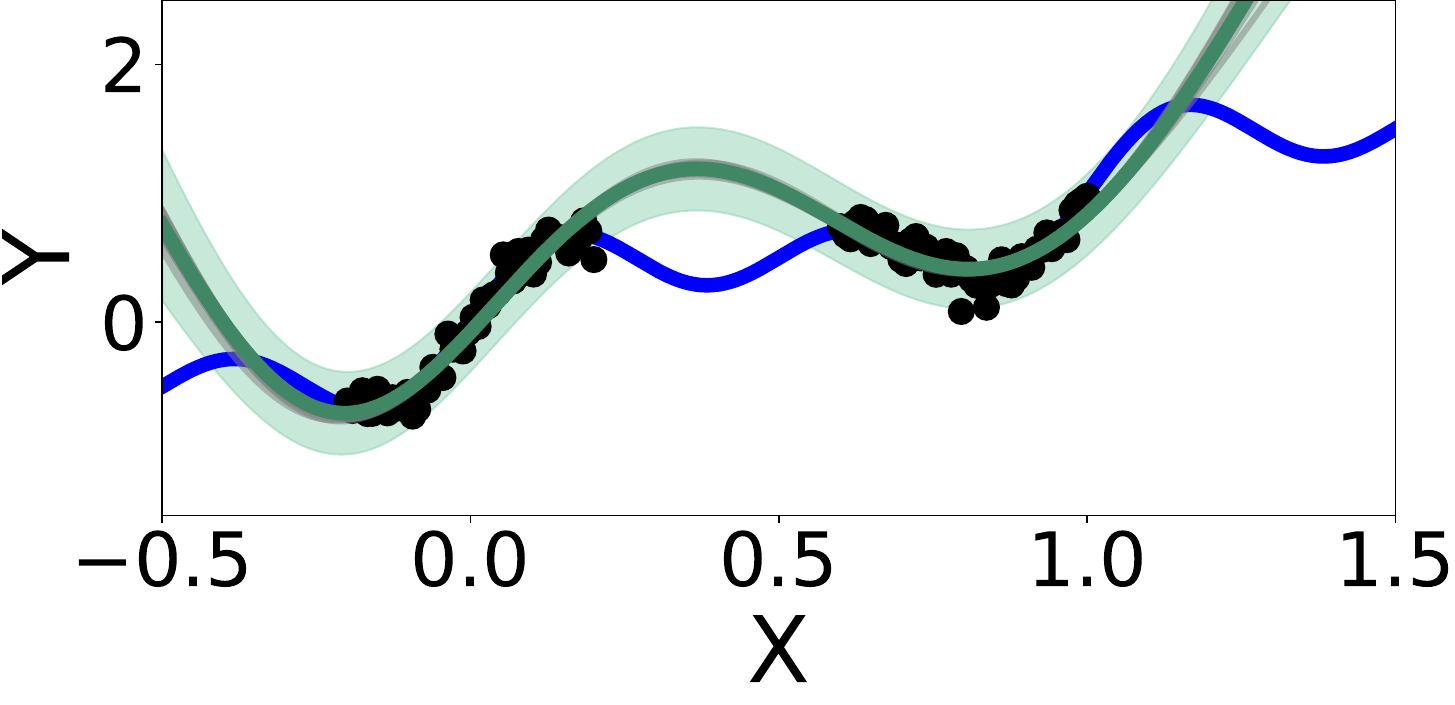}}
\end{minipage}
\begin{minipage}[b]{0.496\linewidth}
    \centering
    \centerline{ \includegraphics[width=\linewidth]{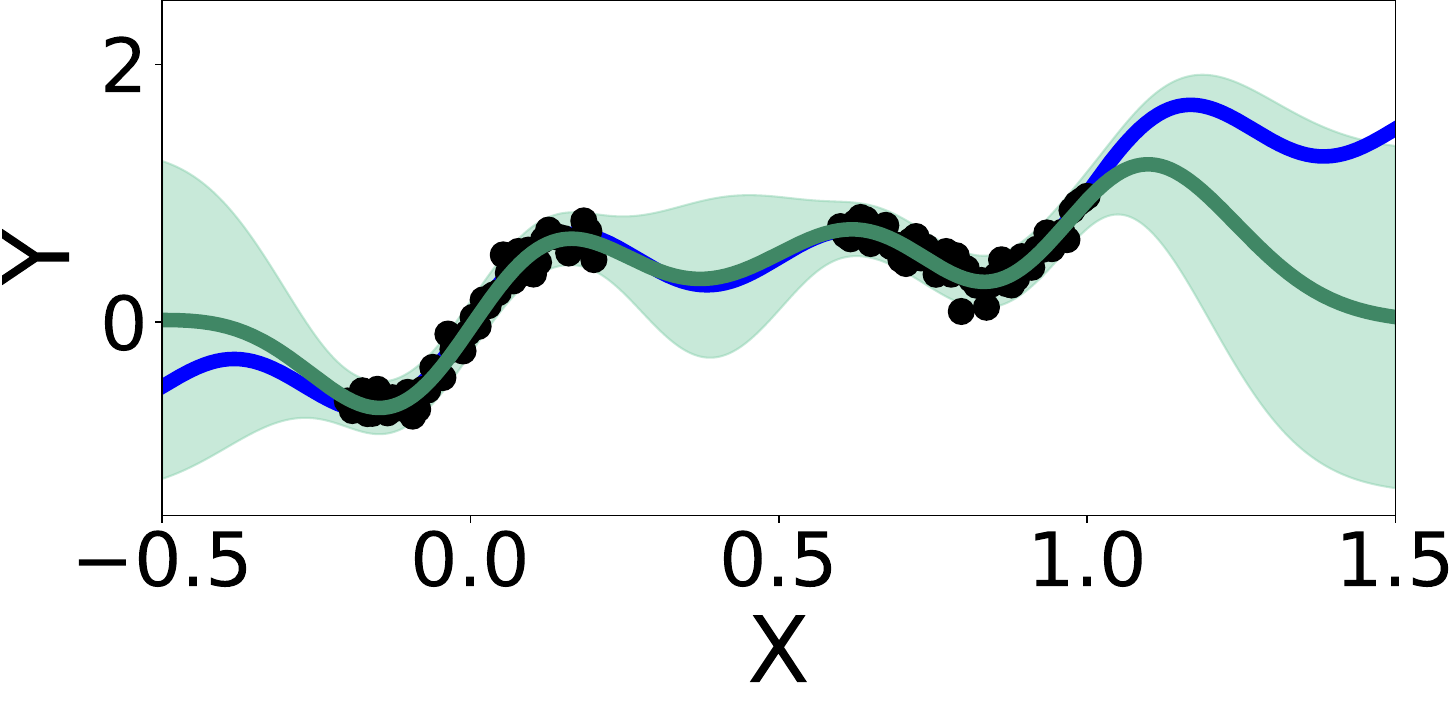}}
\end{minipage}\\
\text{ }\,\, \includegraphics[width=0.97\linewidth]{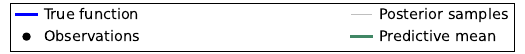}
\vspace{-1em}
\caption{Demonstration of CAS on a univariate regression task.  Top left: Bayesian inference using standard AS \cite{jantre2024active} with $K=20$.  Top right: CAS with an  intermediate dimension $R=250$. Bottom left: CAS on a larger model with $N \approx 4.5 \times 10^{7}$ and $R=10^{4}$. Bottom right: Mean prediction and a $\pm2$ standard deviation band inferred by Gaussian Process Regression.}
\label{fig:univ}
\end{figure}

\begin{table*}[t]
\centering
\caption{Test RMSE / NLL across UCI datasets. Lower is better.}
\vspace{0.2mm}
\label{tab:predictive}
\begin{tabular}{lccc}
\toprule
Dataset & Sketch + Haar & Standard AS & CAS \\
\midrule
Airfoil &  
$4.94\!\pm\!\textcolor{gray}{0.38}$ / ${3.02}\!\pm\!\textcolor{gray}{0.07}$ & 
$4.96\!\pm\!\textcolor{gray}{0.39}$ / ${3.02}\!\pm\!\textcolor{gray}{0.08}$ & 
${4.93}\!\pm\!\textcolor{gray}{0.39}$ / ${3.02}\!\pm\!\textcolor{gray}{0.08}$ \\
Boston &
$4.66\!\pm\!\textcolor{gray}{0.88}$ / $3.00\!\pm\!\textcolor{gray}{0.13}$ &
$4.50\!\pm\!\textcolor{gray}{0.65}$ / $2.97\!\pm\!\textcolor{gray}{0.09}$ &
${4.44}\!\pm\!\textcolor{gray}{0.77}$ / ${2.96}\!\pm\!\textcolor{gray}{0.10}$ \\
Concrete &
$9.90\!\pm\!\textcolor{gray}{0.63}$ / ${3.72}\!\pm\!\textcolor{gray}{0.05}$ &
$9.97\!\pm\!\textcolor{gray}{0.79}$ / ${3.72}\!\pm\!\textcolor{gray}{0.07}$ &
${9.89}\!\pm\!\textcolor{gray}{0.72}$ / ${3.72}\!\pm\!\textcolor{gray}{0.06}$ \\
Energy &
$2.99\!\pm\!\textcolor{gray}{0.29}$ / $2.53\!\pm\!\textcolor{gray}{0.08}$ &
$2.88\!\pm\!\textcolor{gray}{0.23}$ / $2.50\!\pm\!\textcolor{gray}{0.06}$ &
${2.85}\!\pm\!\textcolor{gray}{0.25}$ / ${2.49}\!\pm\!\textcolor{gray}{0.06}$ \\
Yacht &
$9.21\!\pm\!\textcolor{gray}{2.53}$ / $3.75\!\pm\!\textcolor{gray}{0.16}$ &
$8.53\!\pm\!\textcolor{gray}{1.67}$ / ${3.70}\!\pm\!\textcolor{gray}{0.08}$ &
${8.43}\!\pm\!1.85$ / ${3.70}\!\pm\!\textcolor{gray}{0.10}$ \\
\bottomrule
\end{tabular}
\vspace{-1em}
\end{table*}

\begin{table}
\centering
\caption{Empirical coverage of nominal 95\% predictive intervals on UCI datasets.}
\vspace{0.2mm}
\label{tab:coverage}
\begin{tabular}{lccc}
\toprule
Dataset & Sketch + Haar & Standard AS & CAS \\
\midrule
Airfoil & ${0.95}\!\pm\!\textcolor{gray}{0.02}$ & ${0.95}\!\pm\!\textcolor{gray}{0.02}$ & ${0.95}\!\pm\!\textcolor{gray}{0.02}$ \\
Boston & ${0.97}\!\pm\!\textcolor{gray}{0.02}$ & $0.98\!\pm\!\textcolor{gray}{0.02}$ & $0.98\!\pm\!\textcolor{gray}{0.02}$ \\
Concrete & $0.97\!\pm\!\textcolor{gray}{0.02}$ & ${0.96}\!\pm\!\textcolor{gray}{0.02}$ & ${0.96}\!\pm\!\textcolor{gray}{0.02}$ \\
Energy & $0.91\!\pm\!\textcolor{gray}{0.04}$ & ${0.95}\!\pm\!\textcolor{gray}{0.02}$ & $0.94\!\pm\!\textcolor{gray}{0.03}$ \\
Yacht & ${0.96}\!\pm\!\textcolor{gray}{0.03}$ & $0.98\!\pm\!\textcolor{gray}{0.03}$ & $0.98\!\pm\!\textcolor{gray}{0.03}$ \\
\bottomrule
\end{tabular}
\vspace{-0.25em}
\end{table}

\vspace{-0.05in}
\subsection{Univariate regression}
\vspace{-0.03in}
We first demonstrate CAS on the univariate regression task in Fig.~\ref{fig:univ}. Data is generated according to $y = x + 0.3\sin (2\pi x) + 0.3\sin(4\pi x) + \epsilon$, where $\epsilon \sim \mc{N}(0,\sigma_{\epsilon}^2)$ and $\sigma_{\epsilon}=0.1$. We sample 100 values of $x$ in $[-0.2,1]$ with observations omitted in $(0.2,0.6)$ and use $M=500$ gradient samples for subspace construction. We find that CAS closely matches the uncertainty estimates produced by standard AS when both are applied to the same model. In particular, both methods capture increased uncertainty in the region without observations. CAS also scales to the larger model with $N\approx4.5\times10^7$ parameters and $R=10^4$, while continuing to capture meaningful predictive uncertainty.
\vspace{-0.9em}
\subsection{UCI regression tasks}
\vspace{-0.4em}
We evaluate CAS on a suite of UCI regression datasets and compare it with standard AS and a random subspace baseline.
For the latter, $\boldsymbol{P}$ is used to compress the parameter space and a $K$-dimensional subspace is sampled from the Haar measure on orthogonal matrices. We denote this method Sketch + Haar.
We use $K=16$ for all methods. The hidden layer of the neural network model contains $\num{1000}$ units, resulting in model sizes from approximately $N=\num{7000}$ to $N=\num{15000}$ depending on the input dimension. For a given input $\boldsymbol{x}$, we use a Gaussian observation model, and predictive means  $\mu_{\boldsymbol{\theta}}(\boldsymbol{x})$ and variances  $\sigma_{\boldsymbol{\theta}}^2(\boldsymbol{x})$ are estimated from MCMC samples. For CAS and Sketch + Haar, we use $R=256$.

Table~\ref{tab:predictive} reports test root-mean square error (RMSE) and  negative log-likelihood (NLL). CAS closely matches      standard AS across all datasets despite operating in the compressed parameter space. Compared with Sketch + Haar, CAS provides similar or improved predictive performance, with improvements on \emph{Boston}, \emph{Energy}, and \emph{Yacht} datasets. Table~\ref{tab:coverage} shows that CAS also maintains empirical coverage comparable to standard AS and generally close to the nominal $95\%$ level. These results indicate that active subspace construction identifies informative directions within the compressed parameter space compared with selecting a random subspace. Entries in both tables report mean $\pm$ standard deviation across five random seeds and data splits.

\vspace{-0.13in}
\subsection{Scaling behavior}
\begin{figure}
    \includegraphics[width=0.49\linewidth]{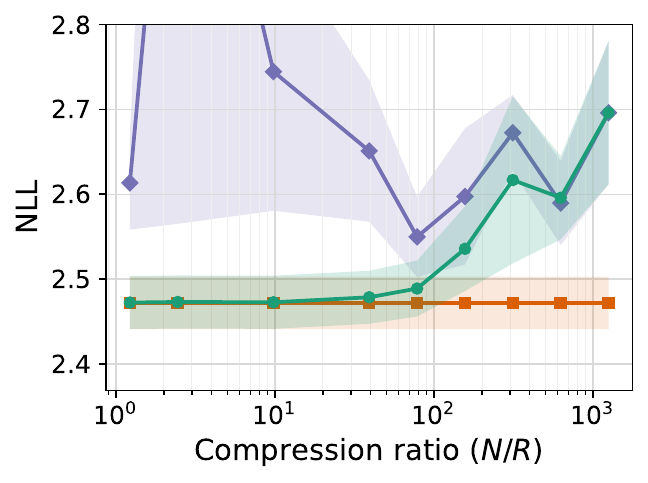}
    \includegraphics[width=0.49\linewidth]{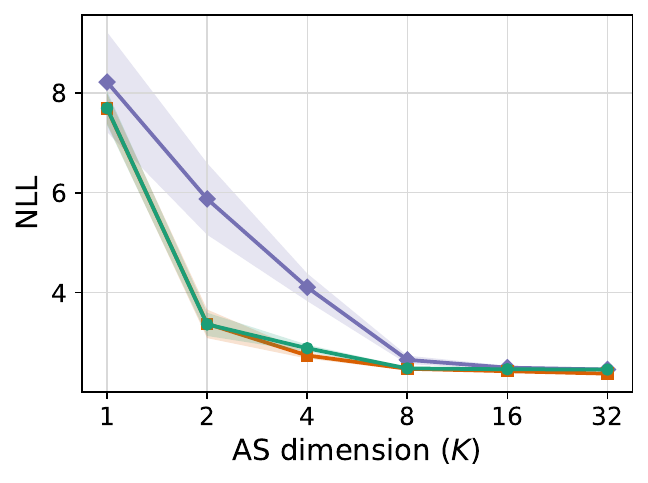}\\
\text{\quad} \includegraphics[width=0.95\linewidth]{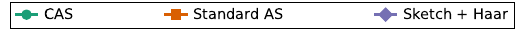}
    \vspace{-1.7em}
    \caption{Left: NLL as the compression ratio varies from $1.2$ (almost no compression) to $\num{1250}$ with $K=8$ fixed. Right: NLL as $K$ varies from $1$ to $32$ with $R=256$ fixed.}
    \label{fig:scaling}
\end{figure}

\vspace{-0.06in}
Here, we study the sensitivity of CAS performance to the intermediate dimension $R$ and active subspace dimension $K$. As a representative dataset, we use the UCI \emph{Energy} dataset with a model containing approximately $10^4$ parameters. Fig.~\ref{fig:scaling} shows that CAS remains competitive with standard AS over a wide range of compression ratios. We also observe that increasing $K$ with $R$ fixed improves the NLL before performance stabilizes at larger active subspace dimensions. For comparison, at significant compression ratios, the random subspaces used in the Sketch + Haar approach are unable to achieve a competitive NLL.

\vspace{-0.05in}
 \subsection{Resource utilization}
\begin{figure}
    \centering
    \includegraphics[width=\linewidth]{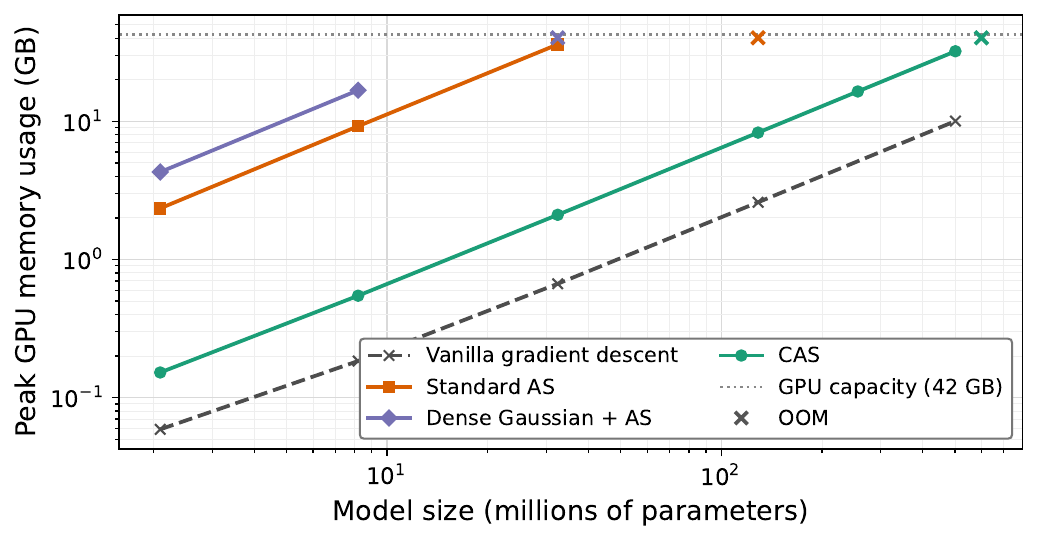}
    \vspace{-2em}
    \caption{Memory usage across model sizes. An $\times$ indicates an out-of-memory (OOM) error. We use $K=16$, $M=128$, and $R=256$ for compressed methods. Baseline GPU memory for a single model gradient is also shown.}
    \label{fig:memory}
\end{figure}

 \vspace{-0.03in}
CAS constructs the active subspace in a reduced space of dimension $R$ instead of the original model parameter space. This allows active subspace construction for significantly larger models, as shown in Fig.~\ref{fig:memory}. Standard AS encounters memory limitations for models above $\approx3\times10^7$ parameters, while CAS scales to $5 \times 10^8$ parameters, more than an order of magnitude larger. The memory-efficient structure of $\boldsymbol{P}$ is crucial for this scaling. Using a dense Gaussian embedding results in memory exhaustion at $\approx3\times10^7$ parameters.

\vspace{-0.1in}
\section{Conclusion and Discussion}
\label{sec:conclusion}
\vspace{-0.05in}
We propose Compressed Active Subspaces (CAS), a scalable approach for predictive uncertainty estimation in high-dimensional models. CAS uses a structured random embedding based on CountSketch matrices to construct the active subspace in a lower-dimensional parameter space. Experimentally, CAS provides predictive performance and uncertainty estimates comparable to standard AS while substantially reducing the dimension required for AS construction. In our experiments, this reduction enables active subspace construction for models with up to $5\times10^8$ parameters, where standard AS becomes computationally prohibitive.

 \vspace{-0.2in}

\section*{Acknowledgment}
\vspace{-0.1in}
This work was supported by the U.S. Department of Energy,
Office of Science, Advanced Scientific Computing Research,
under Contracts DE-SC0012704 and DE-AC02-06CH11357. This research used National Energy Research Scientific Computing Center (NERSC) resources under NERSC awards ASCR-ERCAP 35972 and ASCR-ERCAP 34765.

\end{document}